\documentclass[sigconf,natbib=false]{acmart}
\usepackage[ruled,linesnumbered,vlined]{algorithm2e}
\usepackage{physics}
\AtBeginDocument{%
  }
\allowdisplaybreaks
\definecolor{mygreen}{rgb}{0.0, 0.5, 0.0}
\definecolor{myblue}{rgb}{0.1,0.2,0.6}

\copyrightyear{2022} 
\acmYear{2022} 
\setcopyright{acmlicensed}\acmConference[ICAIF '22]{3rd ACM International Conference on AI in Finance}{November 2--4, 2022}{New York, NY, USA}
\acmBooktitle{3rd ACM International Conference on AI in Finance (ICAIF '22), November 2--4, 2022, New York, NY, USA}
\acmPrice{15.00}
\acmDOI{10.1145/3533271.3561670}
\acmISBN{978-1-4503-9376-8/22/10}

\RequirePackage[
  datamodel=acmdatamodel,
  style=acmnumeric,
  ]{biblatex}

\begin{document}

\title{Optimal Stopping with Gaussian Processes}

\author{Kshama Dwarakanath}
\affiliation{%
  \institution{J.P.Morgan AI Research}
  \city{Palo Alto}
  \state{California}
  \country{USA}
}
\email{kshama.dwarakanath@jpmorgan.com}

\author{Danial Dervovic}
\affiliation{%
  \institution{J.P.Morgan AI Research}
  \city{London}
  \country{UK}}
\email{danial.dervovic@jpmorgan.com}

\author{Peyman Tavallali}
\affiliation{%
  \institution{J.P.Morgan AI Services \& Innovation}
  \city{Palo Alto}
  \state{California}
  \country{USA}}

\author{Svitlana S Vyetrenko}
\affiliation{%
  \institution{J.P.Morgan AI Research}
  \city{Palo Alto}
  \state{California}
  \country{USA}
}

\author{Tucker Balch}
\affiliation{%
  \institution{J.P.Morgan AI Research}
  \city{New York}
  \state{New York}
  \country{USA}
}

\renewcommand{\shortauthors}{Dwarakanath et al.}

\begin{abstract}
We propose a novel group of Gaussian Process based algorithms for fast approximate optimal stopping of time series with specific applications to financial markets. We show that structural properties commonly exhibited by financial time series (e.g., the tendency to mean-revert) allow the use of Gaussian and Deep Gaussian Process models that further enable us to analytically evaluate optimal stopping value functions and policies. We additionally quantify uncertainty in the value function by propagating the price model through the optimal stopping analysis. We compare and contrast our proposed methods against a sampling-based method, as well as a deep learning based benchmark that is currently considered the state-of-the-art in the literature. We show that our family of algorithms outperforms benchmarks on three historical time series datasets that include intra-day and end-of-day equity stock prices as well as the daily US treasury yield curve rates. 
\end{abstract}


\maketitle

\section{Background and Related Work}
The question of when to profitably buy or sell an asset is a long-studied fundamental economic problem~\cite{Rothschild1974, Rosenfield1983}, one that is closely related to the mathematical problem of \emph{optimal stopping}. While optimal stopping problems have a rich literature ~\cite{Wald:1947,Shiryaev2007}, they have historically been solved analytically in restrictive settings where underlying quantities are independent and identically distributed (iid)~\cite{Bruss2000,ferguson2006optimal}, with a known data-generating process. 

Financial price time series such as stock prices display interesting structural properties such as mean-reversion that allow for explicit modeling \cite{poterba1988mean}. Functional data analysis has long been used in modeling time series enabling long term predictions with the ability to work with irregularly sampled data \cite{bengio_gaussian}. In time series modeling, approaches based on Gaussian Processes (GPs) allow long term forecasting in settings with small quantities of data for calibration and those with a need to estimate the covariance of predictions~\cite{Gaussian_time_series,pmlr-v48-hwangb16}. GPs also come up in finance when studying mean reverting processes called Ornstein-Uhlenbeck (OU) processes which are GPs with an exponential kernel \cite{gpml}. In \cite{Duvenaud_thesis}, the author develops a range of methods for automatically choosing GP kernels based on structure observed in the data during training (processes learnt this way are called Deep GPs). In this work, we leverage the effectiveness of the functional form of GPs to provide an efficient solution to optimal stopping problems under minimal assumptions. Such general stopping problems are ubiquitous in finance, for example in the pricing of American options~\cite{McDonald86,Shiryaev2007}.




\paragraph{Contributions.} This work provides the following:
\begin{enumerate}
    \item A novel formulation of the problem of optimally stopping a GP and development of non-adaptive (GPOS, Algorithm~\ref{alg:gpos}) and adaptive (A-GPOS, Algorithm~\ref{alg:agpos}) algorithms to efficiently solve this problem, with allowances for large quantities of data via time series clustering.
    \item Extension of our proposed algorithms to use Deep GPs, which have an advantage of not requiring the specification of a GP kernel beforehand.
    \item Uncertainty quantification for the proposed optimal stopping value functions.
    \item Empirical demonstrations of the effectiveness of our GP-based approach on historical data, namely, intra-day and end-of-day US stock prices and treasury yield curve rates, as well as on synthetic data drawn from an OU process. We demonstrate that our group of algorithms, while being more computationally efficient (exhibiting faster training times) also largely outperforms competing approaches in terms of optimality of the stopping rule.
\end{enumerate}

\paragraph{Related Work.}

From the financial literature, in~\cite{dourban2017optimal}, the authors solve the finite horizon optimal purchasing problem for items with a mean reverting price process, under the assumption the price series follows an AR(1)/discrete OU-process with known parameters and a convex holding cost.
The work in~\cite{lipton2020closedform} presents a closed form solution to this problem. In~\cite{Leung2015,Luu2018}, the authors study optimal trading of a mean-reverting asset, under the assumption that the asset follows an OU process in continuous-time with infinite-horizon.
In the classic work by~\cite{buy_and_hold} on selling an asset following the Black-Scholes model, the optimal strategy is found to either sell immediately or wait until the end of the time horizon, depending on a ``goodness'' measure computable from the model parameters.
A Bayesian optimal selling problem is solved in~\cite{Ming2013} where offers come from one of a finite number of known distributions.
\cite{Sofronov2020} tackles the setting of selling multiple identical assets against a background of sequential iid offer prices and known transaction costs.

From the machine learning literature,~\cite{Tsitsiklis1999} tackle the problem of optimal stopping using Q-learning.
The chosen domain is financial, dealing with pricing a synthetic multidimensional option.
Work using ``Zap'' Q-learning by~\cite{Chen2019} builds on this. The same experimental example is used and compared with the fixed point Kalman filter algorithm of~\cite{Choi2006}.
The work by~\cite{dos} views optimal stopping through the lens of deep learning, using backwards induction to fit a neural network at each timestep to approximate the true value function.
This work is extended to high dimension problems in~\cite{becker_cheridito_jentzen_welti_2021}.

Optimal Stopping can be formulated~\cite{puterman2014markov} as a Markov Decision Process (MDP) with time in the state space and with a highly restrictive action space.
This specialised structure of the optimal stopping MDP renders the problem unsuitable for use with general MDP solvers, due to data sparsity and lack of influence of the decision-maker on the state transition process.
In~\cite{DeisenrothACC, NIPS2003_7993e112}, the authors address solving general MDPs using GPs.
For~\cite{NIPS2003_7993e112} the MDP dynamics and the value function are learned as GPs using policy iteration at finite number of ``support points''. For optimal stopping problems the required number of support points would be prohibitively large, $O(W T)$ with $W$ being number of discretization points in ``price'' space and $T$ the time horizon.
The GPDP algorithm presented in~\cite{DeisenrothACC} would engender a similar computational burden when applied to the optimal stopping setting, with a distinct GP being trained for the value function at each time-step, and a $Q$-function for each (training input, time-step) pair.

The closest work to the present paper is~\cite{pmlr-v97-dai19a}, wherein the authors solve a hyperparameter optimisation problem by using a GP within a Bayesian Optimal Stopping scheme originally defined in~\cite{MULLER20073140} to prune poorly performing hyperparameter configurations while training the corresponding neural network.
This work uses a GP with a specific kernel (enforcing monotonicity) with a value function approximation based on Monte Carlo sampling. The GP based family of optimal stopping algorithms presented here uses a piecewise constant value-function approximation evaluated using Gaussian integrals to analytically estimate stopping value functions in a tractable fashion, with the underlying time-series being modelled by a GP. We further use Deep GPs to generalize the proposed algorithms when the GP kernel cannot be specified in advance.

\subsection{Optimal Stopping}\label{subsec:optimal_stopping}
Following the notation in \cite{ferguson2006optimal}, stopping rule problems are defined by two objects: a sequence of random variables $X_1, X_2, \cdots$ whose joint distribution is assumed known, and a sequence of real-valued reward functions $y_0,\ y_1(x_1),\ \cdots,\  y_\infty(x_1, x_2,\cdots)$. Given these two objects, the stopping rule problem can be described as follows. You observe the sequence $X_1, X_2, \cdots$ for as long as you wish. For each $n = 1, 2, \cdots$, after observing $X_1 = x_1,\ X_2 = x_2,\ \cdots, X_n = x_n$, you may stop and receive the known reward $y_n(x_1, x_2,\cdots,x_n)$, or you may continue and observe $X_{n+1}$. If you choose to continue (and not take any observations), you receive the constant amount, $y_0$. If you never stop, you receive $y_\infty(x_1, x_2,\cdots)$.

The goal is to choose a time to stop to maximise the expected reward. Let $\phi_n(x_1, \cdots,x_n)\in\lbrace0,1\rbrace$ denote the (deterministic) stopping decision made at stage $n$ having observed $X_1 = x_1,\ \cdots, X_n = x_n$, where $\phi_n(x_1,\cdots,x_n)=1$ denotes the decision to stop, and $\phi_n(x_1,\cdots,x_n)=0$ denotes the decision to continue. The observations $(X_1, X_2, \cdots)$ and sequence of decisions $\phi=\lbrace\phi_n(X_1,\cdots,X_n):n\geq0\rbrace$ determine the random time $N$ at which stopping occurs, where $0 \leq N \leq \infty$ and $N=\infty$ means stopping never occurs. Our task is then to choose a stopping rule $\phi$ to maximise the expected return, defined as\begin{align}
    V(\phi) &= \mathbb{E}\left[y_N(X_1, \cdots, X_N)\right]\nonumber
\end{align}
We are interested in stopping rule problems with a finite horizon, those that require stopping after observing $X_1 , \cdots , X_T$ for some $T<\infty$. They are obtained as a special case of the general problem above by setting $y_{T+1} = \cdots = y_\infty = -\infty$.

In this work, the random variables $X_1,\cdots,X_T$ represent asset prices at times instants $1,\cdots,T$. For the case where we want to find the best time and price at which to sell the asset, the reward function $y_t(x_1,\cdots,x_t)=x_t$ represents the asset price at time $t$. Then, the goal of the stopping problem is to find the maximum expected (selling) price and the time of reaching that price. For the case where we want to find the best time and price to buy the asset, the reward function $y_t(x_1,\cdots,x_t)=-x_t$ represents the negative of the asset price at time $t$. Then, the goal of the stopping problem is to find the minimum expected (buying) price and the time of reaching that price. For the sake of clarity, we develop our technique for the former case of finding the best time and price at which to sell an asset. Since the selling decision is made at a single time step, we ignore transaction costs of asset trading in this work.
Note that our developed methods are capable of solving any optimal stopping problem even though the discussion in this paper is focused on finding the best time to sell (or buy) an asset. This would require passing the GP model for random variables $X_i$ through the reward functions $y_i(\cdot)$ while computing the stopping value functions.
\subsection{Dynamic Programming}

In principle, finite horizon stopping problems may be solved by the method of backward induction. Since we must stop at time $T$, we first find the optimal rule at time $T - 1$. Upon knowing the optimal rule at $T - 1$, we find the optimal rule at time $T - 2$, and so on back to the initial time $0$ using the \emph{Principle of Optimality} \cite{bellman1954theory}. Define $V_t(x_1,\cdots,x_t)$ to be the maximum reward one can obtain starting from time $t$ after having observed $X_1 = x_1 , \cdots , X_t = x_t$. Then, $V_T(x_1, \cdots, x_T) = y_T(x_1, \cdots, x_T)$, and inductively for $t = T - 1,T-2,\cdots,0$:
\begin{equation}
    \begin{aligned}
    &V_t(x_1, \cdots, x_t)=\max\Big\lbrace y_t(x_1,\cdots,x_t),\\
    &\mathbb{E}\Big[V_{t+1}(X_1, \cdots, X_t, X_{t+1})\Big|X_1 = x_1, \cdots, X_t = x_t\Big]\Big\rbrace
    \end{aligned}\label{eq:value_fn_dp}
\end{equation}
We compare the reward for stopping at time $t$ namely $y_t(x_1,\cdots , x_t)$, with the reward we expect to get by continuing and using the optimal rule for stages $t + 1$ through $T$. Our optimal reward is therefore the maximum of these two quantities. And, it is optimal to stop at $t$ if $\mathbb{E}\Big[V_{t+1}(X_1, \cdots, X_t, X_{t+1})\Big|X_1 = x_1, \cdots, X_t = x_t\Big] \leq y_t(x_1, \cdots ,x_t)$, and to continue otherwise. The value of the stopping rule problem is $V_0$. 

One can see that even for finite horizon stopping problems with a known distribution over $(X_1, \cdots, X_T)$, finding a solution is non-trivial with the main difficulty being the determination of the value functions (\ref{eq:value_fn_dp}). This bottleneck can be alleviated by the assumption of a distributional form for the reward function $y_t(x_1,\cdots,x_t)$ as in the next subsection. 

\subsection{Gaussian Processes}
In this work, we model the asset price (consequently the stopping reward function) as a function of time with a Gaussian Process (GP) prior.
A stochastic process $f(s)$ is said to be Gaussian if and only if for every finite set of indices $n$, the variable $(f(s_1), \cdots , f(s_n))$ is a multivariate Gaussian random variable \cite{gpml}. A GP is entirely specified by its mean function $\mu(s)$, and its covariance function $k(s, s') = \operatorname{Cov}[f(s), f(s')]$. 
The covariance function or kernel $k(\cdot,\cdot)$ is symmetric and positive definite. By convention, the mean function $\mu(s) = 0$ and we say $f \sim \mathcal{GP}(0, k(\,\cdot\, , \,\cdot\, ))$. We also assume a Gaussian likelihood for the asset price $y_t(x_1,\cdots,x_t)=x_t$, that is, $p(y \mid f) \sim \mathcal{N}(f, \sigma_n^2)$, where $\sigma_n^2$ is a (small) variance.
Equivalently, $y_t(x_1,\cdots,x_t) = x_t = f(t) + \epsilon$ where $\epsilon$ is Gaussian noise with variance $\sigma_n^2$.

Given training data $\mathcal{D}=\lbrace y_1,\cdots,y_n\rbrace$ comprising asset prices at time instants $1,\cdots,n$, we infer the distribution of $f(s_*)$ at a new time instant $s_*$ as follows. Let $\vb{y}=\begin{bmatrix}y_1&\cdots&y_n\end{bmatrix}^\top$ be the vector of observed values, $K$ be the kernel matrix with elements $[K]_{i,j} = k(s_i, s_j)$ and $\vb{k}_*$ be the vector with elements $[k_*]_{i} = k(s_*, s_i)$. Then, one can show that the distribution of $f(s_*)$ at a new time instant $s_*$ is Normal, with mean and variance given by
\begin{align}
    \mathbb{E}[f(s_*)] &= \vb{k}_*^\top (K + \sigma^2_n I)^{-1} \vb{y}, \label{eq:gp_mean}\\
    \operatorname{Var}[f(s_*)] &= k(s_*, s_*) - \vb{k}_*^\top (K + \sigma^2_n I)^{-1} \vb{k}_*, \label{eq:gp_variance}
\end{align}
The presence of the matrix inverse in (\ref{eq:gp_mean})-(\ref{eq:gp_variance}) results in a computation time of $O(n^3)$ that can be arduous for applications with $n\geq10^4$. We bypass this limitation of GPs in this work by clustering samples in the training set, and fitting a GP to the centroid (most representative sample) of each cluster. The computational complexity is further reduced by clustering an initial warm start period of data. 


\subsection{Deep Gaussian Processes}

While GPs are attractive for modeling structured time series data using an appropriate kernel, the quality of their fit depends on the knowledge of this structure beforehand. For mean reverting time series, the fact that the series shows mean reversion is used to fit a GP with an exponential kernel. In cases when such structural information is not known beforehand, the common approach is to fit multiple GPs each with a different kernel and to pick one that performs best on a validation set. On the other hand, deep Gaussian Processes (denoted DGPs henceforth) constitute a model for time series where the kernel is learned from the data as described in Chapter 5 of \cite{Duvenaud_thesis}. DGPs model a distribution on functions constructed by composing functions drawn from GP priors. They are effectively a \emph{deep} version of regular GPs where the outputs of a vector of GPs is fed into the inputs of the next vector of GPs \cite{damianou2013deepgp}. One can think of a DGP as a neural network with multiple hidden layers with each hidden unit constituting a GP with a predefined kernel, such as a Radial Basis Function (RBF), rational quadratic or linear kernel \cite{kern_cook}. The weights of the DGP are fit (typically) using variational inference using the training data. 

\begin{figure}[tb]
    \centering
    \includegraphics[width=\linewidth,height=1.8in]{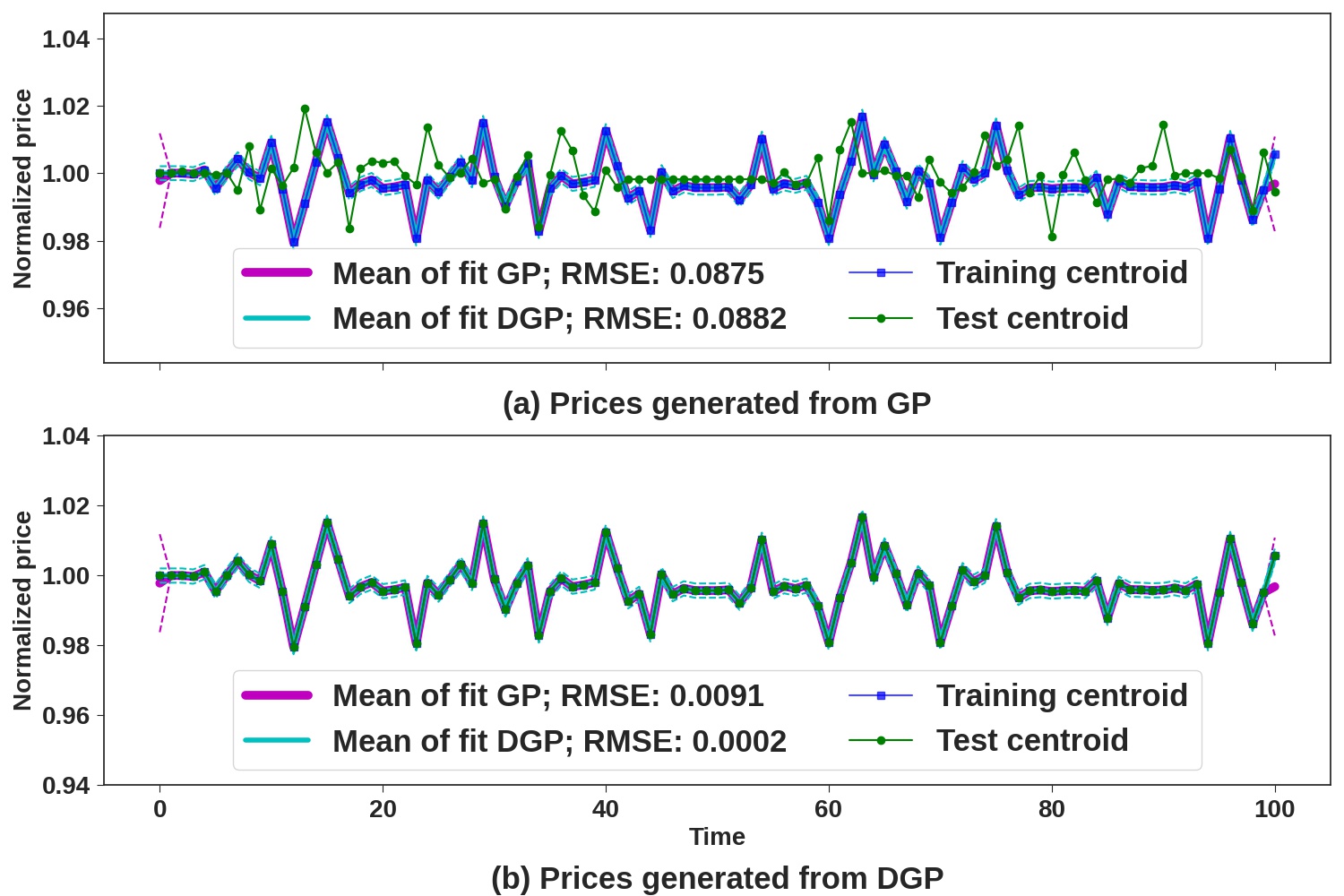}
    \caption{Motivation for using DGPs}
    \label{fig:dgp_on_gp}
\end{figure}

The main advantage of DGPs over GPs for modeling structured time series is that the exact knowledge of the structural form is not as necessary for DGPs as it is for GPs. Loosely speaking, the \emph{effective} kernel in DGPs is learnt from data as opposed to being pre-specified for GPs. To illustrate this, we generate two sets of synthetic asset price data and partition each into training and test sets. The first set is generated from a GP with an exponential kernel. We fit a GP and DGP to the most representative sample (or centroid) of the training set and examine the GP mean function and root mean squared error (RMSE) in prediction over the test set in Figure \ref{fig:dgp_on_gp} (a). 
We see that the RMSE in predictions over the test set for both fits are fairly close to each other. For the second data set, we generate data from a DGP with a single hidden layer. We compare the RMSE in predictions for a GP and a DGP fit to the centroid of the training set in Figure \ref{fig:dgp_on_gp} (b). In this case, we see that while the RMSE of the fit DGP is small, that for the fit GP is much larger. This shows that a DGP model (when trained properly\footnote{To ensure proper training of the DGP models, we made sure their predictive performance on the training data set improved during training.}) does similarly or better than a GP model fit to the same data. Moreover, a GP model may not have the capacity to fit data generated from a DGP model due to its kernel limitations. Thus, DGPs allow for automated kernel fitting while enjoying similar structural properties and computational advantages as GPs. Going ahead, we adopt both models in designing our optimal stopping algorithms. The Deep GP package \cite{dgp_package} is used to construct and train our DGP models.

\section{Problem Setup}\label{sec:problem}
We are interested in forecasting for the best time and price to sell an asset given historical asset prices. Let $\mathcal{D}=\lbrace y^i_t:1\leq i\leq N,1\leq t\leq T\rbrace$ be training data with $y^i_t$ being the asset price at time $t$ in sample $i$. We build upon the idea that intra-day and end-of-day asset prices are mean reverting and hence, model them using a GP. Assuming that this structural information helps us compute the value function in (\ref{eq:value_fn_dp}), we compute the stopping policy as the decision to continue (holding the asset) or to stop (and sell the asset) in (\ref{eq:value_fn_dp}). Given new (test) asset price series $y^*=\lbrace y^*_1,\cdots,y^*_W\rbrace$ up until $W<T$ steps in the day, we wish to use the optimal policy to find the best time and price to sell the asset. 

\subsection{Approximations to the value function}
To make the problem more tractable, we make two assumptions to compute the value function. We first assume that the value function in (\ref{eq:value_fn_dp}) depends only on the asset price at the current time instant and the training data $\mathcal{D}$ as\begin{align}
    V_t(x_1,\cdots,x_t)\approx \hat{V}_{t}(x_t|\mathcal{D})\label{eq:ass1}
\end{align}
This is to say that the maximum asset price going forward from time $t$ is a function of only the asset price at time $t$, and previously observed asset price data. 

The second approximation is to assume that the value function conditioned on the training data $\hat{V}_t(x_t|\mathcal{D})$ is piecewise constant in asset price $x_t$. Partition the real line into $M<\infty$ disjoint intervals $b_i=[b_i^-, b_i^+)$, with $\mathcal{B}=\cup_{i=1}^M[b_i^-,b_i^+)$ being the collection of all such intervals\footnote{We are currently working on renouncing the bins and estimating a continuous stopping value function as a direct extension to this work.}. Note that $b_i^-$ and $b_i^+$ are the lower and upper limits respectively for a bin $b_i \in \mathcal{B}$, and write the bin center as $\bar{b}_i = \frac{b_i^- + b_i^+}{2}$. Then, we have\begin{align}
    \hat{V}_t(x_t|\mathcal{D})&\approx \sum_{i=1}^M\hat{V}_{t,i}(\mathcal{D})\delta_{x_t\in b_i}\label{eq:ass2}
\end{align}
where $\delta_{a\in A}:=\begin{cases}1\textnormal{, if }a\in A\\0\textnormal{, otherwise}\end{cases}$. The approximation in (\ref{eq:ass2}) becomes an exact equality as $M\rightarrow\infty$. \footnote{We found it sufficient to pick bins of equal length to cover the range $\left(0.8\times\min\left(y_t^i\right),1.2\times\max\left(y_t^i\right)\right)$.}


Using assumptions (\ref{eq:ass1}) and (\ref{eq:ass2}) in (\ref{eq:value_fn_dp}), we have\begin{align}
    &\mathbb{E}\Big[V_{t+1}(X_1, \cdots, X_t, X_{t+1})\Big|X_1 = x_1, \cdots, X_t = x_t\Big]\nonumber\\
    &\approx\sum_{j=1}^M\hat{V}_{t+1,j}(\mathcal{D})\int_{b_j^{-}}^{b_j^{+}}\mathbb{P}\Big(X_{t+1}=z\Big|\mathcal{D}\Big)dz\label{eq:value_approx}
\end{align}
If $x_t\in [b_i^-,b_i^+)=b_i$, putting (\ref{eq:value_approx}) back into (\ref{eq:value_fn_dp}) gives\begin{equation}
    \begin{aligned}
    &\hat{V}_{t,i}(\mathcal{D})=\max\left\lbrace \bar{b}_i,\sum_{j=1}^M\hat{V}_{t+1,j}(\mathcal{D})\int_{b_j^{-}}^{b_j^{+}}\mathbb{P}\Big(X_{t+1}=z\Big|\mathcal{D}\Big)dz\right\rbrace
    \end{aligned}\label{eq:gpos_value}
\end{equation}
with the boundary condition $\hat{V}_{T,k}(\mathcal{D})=\bar{b}_k$ where $x_T\in b_k$.

Note that the GP (or DGP) model assumption makes the computation of the integral in (\ref{eq:gpos_value}) simple using Gaussian distribution functions. The corresponding (approximately) optimal policy is given as \begin{equation}
    \hat{\pi}(t,i)=\begin{cases}1\textnormal{, if }\hat{V}_{t,i}(\mathcal{D})=\bar{b}_i\\0\textnormal{, otherwise}\end{cases}\label{eq:gpos_policy}
\end{equation}
where $\hat{\pi}(t,i)=1$ denotes that it is optimal to sell the asset at time $t$ if the current asset price lies in bin $b_i$. Likewise, $\hat{\pi}(t,i)=0$ denotes that it optimal to hold the asset. Note that we have decomposed the problem of finding the best time and price to sell the asset into a series of stopping decisions $\hat{\pi}(\cdot,\cdot)$ specifying the action to take at every time based on the asset price.

\subsection{Metric to measure quality of stopping policy}
Now that we are theoretically able to find a stopping policy as in (\ref{eq:gpos_policy}), we need a metric to quantify its performance. Define the suboptimality of a policy $\hat{\pi}:\lbrace1,\cdots,T\rbrace\times\lbrace1,\cdots,M\rbrace\rightarrow\lbrace0,1\rbrace$ over a price series $y=\lbrace y_1,\cdots,y_T\rbrace$ as \begin{align}
    \mathrm{Sub}(\hat{\pi},y)=\left(\max_{1\leq t\leq T}y_t\right)-\sum_{t=1}^T\sum_{i=1}^M\hat{\pi}(t,i)\cdot y_t\cdot \delta_{y_t\in b_i}\label{eq:suboptimality}
\end{align}
(\ref{eq:suboptimality}) captures the difference in the true maximum asset price, and the predicted maximum price resulting from policy $\hat{\pi}$. Clearly, $\mathrm{Sub}(\hat{\pi},y)\geq0$ with lower values characterising \emph{better} stopping policies. When the prices are given in \$ amounts, $\mathrm{Sub}(\hat{\pi},y)$ as computed above is also in \$ amount. When comparing suboptimality values across assets, it is helpful to express $\mathrm{Sub}(\hat{\pi},y)$ in basis points (denoted bps henceforth). We compute the suboptimality in bps by dividing the suboptimality in \$ by the arithmetic mean of sample asset prices in \$, and multiplying the result by $10^4$.

\subsection{Uncertainty quantification}
One of the main advantages of using a GP as a model for time series is the predictive uncertainty that comes along with it. Using the fact that the price series is modeled as a GP (or DGP), we can express the price random variable as $X_{t}\sim p_{X_{t}}\left(.|\mathcal{D}\right)$. Since we further bin the asset prices, we introduce the following random variable for bins $\left\{ b_{i}\right\} _{i=1}^{M}$:
\begin{align}
    S_t=\sum_{i=1}^M\delta_{X_{t}\in b_{i}}\cdot i\nonumber
\end{align}
This allows us to express the value function for each bin in a probabilistic way paving the way to estimate the uncertainty of the value function. 
If we define $p_{t,j}\coloneqq\mathbb{P}\left(S_{t}=j\right)=\int_{b_j^{-}}^{b_j^{+}}p_{X_{t}}\left(z|\mathcal{D}\right)dz$, and $E_{t+1}=\mathbb{E}_{S_{t+1}}\left[\hat{V}_{t+1,S_{t+1}}\left(\mathcal{D}\right)\right]$, the probability mass function (PMF) of $\hat{V}_{t,S_t}(\mathcal{D})$ is
\begin{align}
\mathbb{P}\left(\hat{V}_{t,S_t}(\mathcal{D})=v\right)=
    \begin{cases}
    0,\textnormal{ if }v<E_{t+1}\\
    \sum_{i:\bar{b}_i<E_{t+1}}p_{t,i},\textnormal{ if }v=E_{t+1}\\
    p_{t,j},\textnormal{ for all }j:v=\bar{b}_{j}>E_{t+1}
    \end{cases}\label{eq:uq_value}
\end{align}
Intuitively, the distribution in (\ref{eq:uq_value}) is a folded version of the binned price distribution as any value smaller than $E_{t+1}$ gets a zero probability. (\ref{eq:uq_value}) can be directly used to evaluate measures of uncertainty (e.g. variance) for the value function.

\begin{figure}[b]
    \centering
    \includegraphics[width=\linewidth]{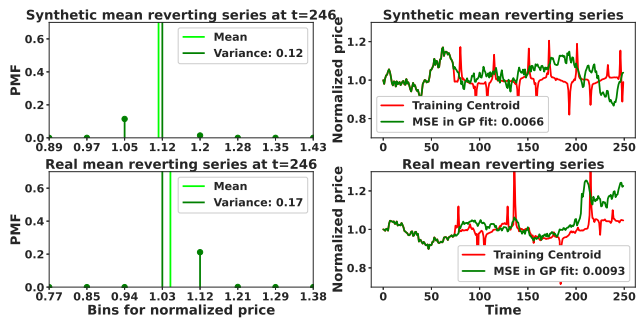}
    \caption{PMF of value function for synthetic and real mean reverting data}
    \label{fig:uq_value}
\end{figure}

Figure \ref{fig:uq_value} is a plot of the PMF of the value function (\ref{eq:uq_value}) at a chosen time $t$, along with the asset prices used to fit the GP model. The top row of plots corresponds to fitting a GP model to synthetically generated asset prices from a GP with an exponential kernel. The bottom row corresponds to fitting a GP model to real asset prices that are seen to be mean reverting. The right column of plots shows the training centroid over which the GP model is fit, and the test sample along with the mean squared error (MSE) in the GP fit over the test sample. Intuitively, one would expect the GP model to be a better fit for the synthetically generated data than the real asset price data. 
And, we see that the quality of GP fit translates into the variance in the value function. 

\section{Approach}\label{sec:algorithms}
Our method works by fitting a GP (or DGP) to pre-processed training data $\mathcal{D}$ to evaluate the term inside the integral in (\ref{eq:gpos_value}). 

\subsection{Time Series Clustering}\label{subsec:clustering}


We preprocess the training set $\mathcal{D}$ by partitioning it into $K\geq1$ clusters as $\mathcal{D}=\mathcal{D}_1\cup\mathcal{D}_2\cdots\cup\mathcal{D}_K$ with each cluster $\mathcal{D}_k$ containing $n_k$ training samples over the initial time period $\lbrace1,\cdots,W\rbrace$. Define the log return (or simply return) of an asset with price $y_t$ at time $t$ as \begin{align}
    r_t=\log y_{t+\delta t} - \log y_t\nonumber
\end{align}
where $\delta t$ is a pre-defined time scale for computing returns. The standard deviation of asset returns is a measure of asset \emph{volatility}. The clustering method takes in log returns of asset price time series as input and clusters them by grouping time series with similar volatility together. The reason for this clustering stems from the fact that asset prices can exhibit different levels of volatility based on market regimes as well as seasonal fluctuations across different assets \cite{getreal}. This also enables us to pick the most relevant cluster to compute our optimal stopping policy when presented with new (test) asset prices. 

We leverage a clustering algorithm called Dynamic Time Warping (DTW) that seeks to find a minimal-distance alignment between times series that is robust to shifts or dilations in time, unlike the classical Euclidean metric \cite{dtw_original,berndt1994dtw}. For a pair of series of lengths $m$ and $n$, there exists an $O(mn)$ algorithm that solves DTW using dynamic programming \cite{tslearn}. 
After partitioning the training set $\mathcal{D}$ into $K$ clusters, we summarize each cluster with a \emph{centroid} time series that is meant to capture the essence of the series in each cluster. We compute the DTW centroids using the DTW Barycenter Averaging Algorithm (DBA) proposed in \cite{dtw_barycenter}. We plug-and-play these time series clustering and centroid computation methods from the Tslearn package \cite{tslearn}. Note that $K=1$ corresponds to summarizing \emph{all} of the data in $\mathcal{D}$ with a single centroid. 

\begin{figure}[b]
    \centering
    \includegraphics[width=\linewidth,height=1.7in]{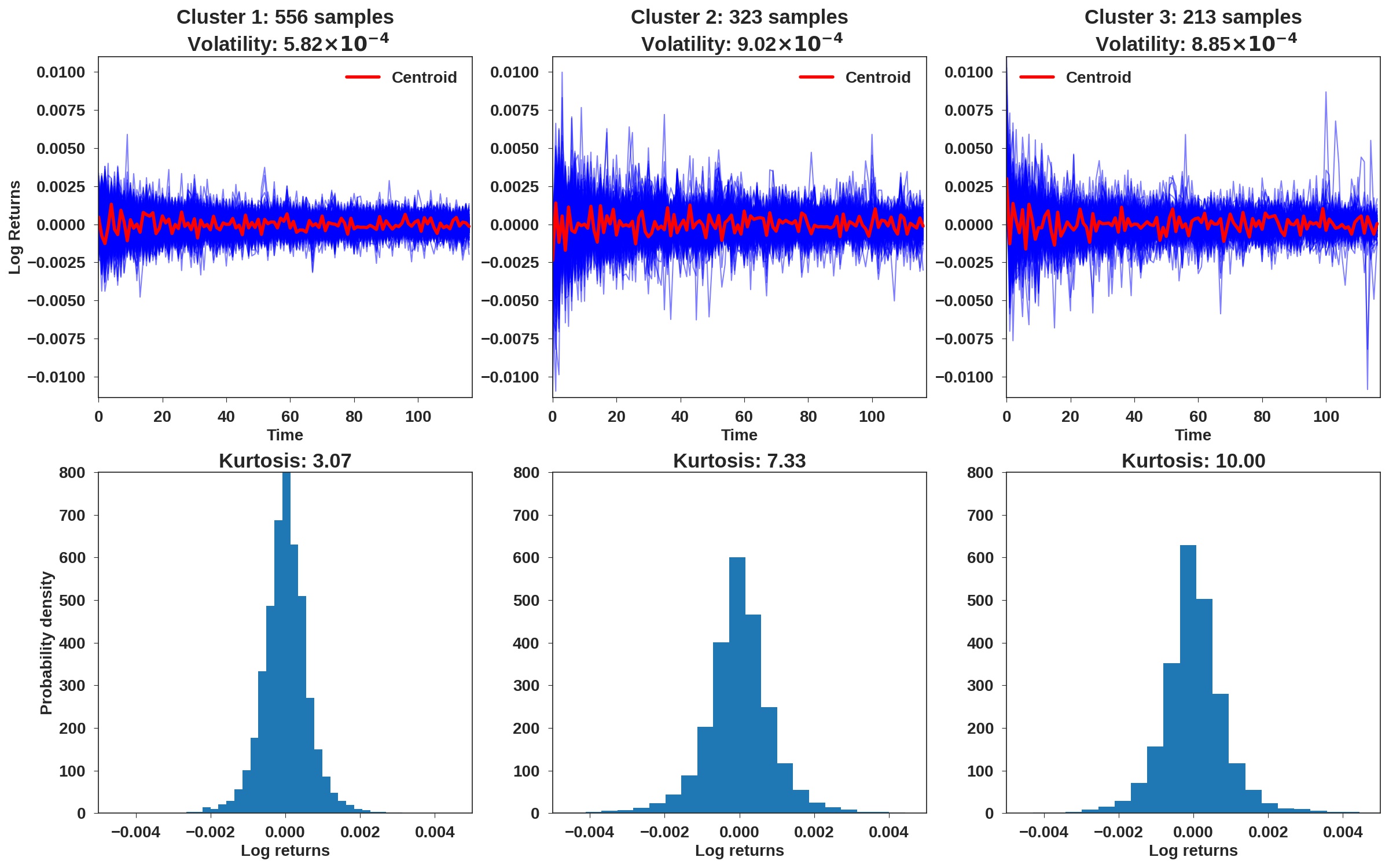}
    \caption{Clusters with differences in volatility and kurtosis}
    \label{fig:clus}
\end{figure}

Figure \ref{fig:clus} shows an example of DTW clustering of log returns with $K=3$ clusters. While the first row plots log returns along with the centroid of each cluster, the second row shows histograms of log returns. We also measure the volatility (or standard deviation) of returns in each cluster, and observe a distinction in volatility between clusters. Another stylized metric of interest for financial time series is the kurtosis or \emph{tailedness} of returns distributions \cite{getreal}. A large kurtosis value signifies higher deviation of the returns distribution from a Gaussian distribution. And, we observe that while cluster 1 has low kurtosis meaning that it is fairly close to a Gaussian distribution, clusters 2 and 3 have higher kurtosis.

\subsection{Algorithms}\label{sec:proposed_algos}

After clustering the training data and computing the centroids of each cluster, we fit a GP (or DGP) to these centroids. This GP (or DGP) fit is used to compute the value function and stopping policy as in (\ref{eq:gpos_value})-(\ref{eq:gpos_policy}) for each cluster. When presented with (test) asset price series $y^*=\lbrace y^*_1,\cdots,y^*_W\rbrace$ up until $W<T$ steps in the day, we first assign a cluster to $y^*$ based on the clustering on the training data. This helps us assign the test series to a cluster with similar statistical properties such as volatility. We then use the stopping policy of that cluster to find the best time and price to sell for the test series. Our algorithm is called Gaussian Process based Optimal Stopping (GPOS) and is presented in Algorithm \ref{alg:gpos}.

Note that GPOS does not change the GP (or DGP) fit after observing the new test series. This can result in poor performance when the test series is rather different from the training set. To handle such cases, we propose an adaptive version of GPOS that re-fits the GP (or DGP) upon observing $W$ time steps of the new test series, while using past information for the remaining time steps of $W+1,\cdots,T$. This is given in Algorithm \ref{alg:agpos} where the computation of the value function and stopping policy are done only after observing the test series, with the training set used for clustering and filling up the centroids after $W$ time steps. 

\subsection{Benchmarks}

\paragraph{Sample Optimal Stopping.}

We propose an intuitive benchmark for optimal stopping algorithms that computes the stopping policy without using any structural information about the time series (such as mean reversion). We call it Sample Optimal Stopping (SOS) since it computes approximations to the value function (\ref{eq:value_fn_dp}) using the asset price samples in set $\mathcal{D}$. Define the value function for sample $i$ at time $t$ to be the highest asset price seen at/after $t$:\begin{align}
    V^i_t=\max_{t\leq s\leq T}y_s^i\label{eq:sos_value}
\end{align}
and the average value function for the set $\mathcal{D}$ at time $t$ as\begin{align}
    V_t=\frac{1}{N}\sum_{i=1}^N V^i_t\label{eq:sos_avg_value}
\end{align}
(\ref{eq:sos_avg_value}) estimates the (true) value function used to compute the stopping policy as $\hat{\pi}(t,i)=\begin{cases}1\textnormal{, if }\bar{b}_i\geq V_t\\0\textnormal{, otherwise}\end{cases}$ when the price lies in bin $b_i$.

\begin{figure}[b]
    \centering
    \includegraphics[width=\linewidth,height=1.7in]{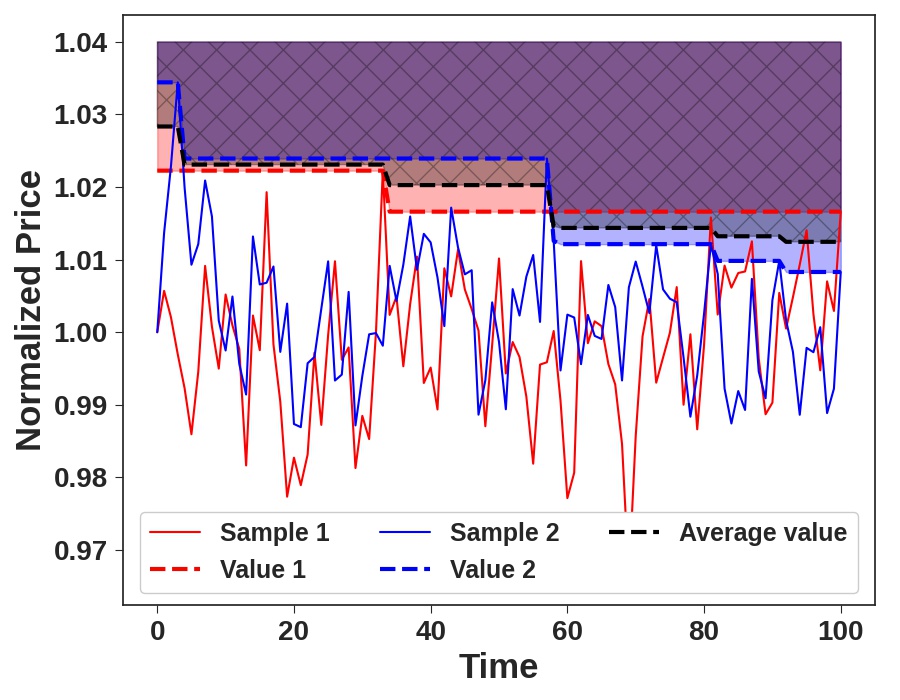}
    \caption{Value function in SOS on two training data samples}
    \label{fig:sos_optimal_policy}
\end{figure}

Figure \ref{fig:sos_optimal_policy} plots (\ref{eq:sos_value}) and (\ref{eq:sos_avg_value}) in dashed lines along with asset prices in solid lines for a training set with $N=2$ samples and time horizon $T=100$. The shaded (or hatched) region above each value function denotes the stopping policy that dictates that the asset be sold upon observing a price higher than the corresponding value. And, be held otherwise. 

\begin{algorithm}[tb]
\caption{GPOS: Gaussian Process based Optimal Stopping\label{alg:gpos}}
\DontPrintSemicolon
\SetKwFunction{Training}{Training}
\SetKwProg{Fn}{Function}{:}{}
  \Fn{\Training{$\mathcal{D}$}}{
    Partition $\mathcal{D}$ into $K\geq1$ clusters $\mathcal{D}_1,\cdots,\mathcal{D}_K$\;
    \For{$k\in\lbrace1,\cdots K\rbrace$}{
    Fit GP or DGP to \emph{centroid} of cluster $k$\;
    Compute $\hat{V}_{t,i}(\mathcal{D}_k)$ as in (\ref{eq:gpos_value}) for all $t,i$\;
    Compute $\hat{\pi}_k(t,i)$ as in (\ref{eq:gpos_policy}) for all $t,i$\;
    }
    \KwRet{$\lbrace\hat{\pi}_1,\cdots,\hat{\pi}_K\rbrace$}\;
  }
\SetKwFunction{Testing}{Testing}
\SetKwProg{Fn}{Function}{:}{}
\Fn{\Testing{$\lbrace\hat{\pi}_1,\cdots,\hat{\pi}_K\rbrace$, $y^*$}}{
Find cluster $k$ for $y^*$\;
\For{$t\in\lbrace W+1,\cdots,T\rbrace$}{
Find bin $b_i=[b_i^-,b_i^+]\ni y^*_t$\;
\eIf{$\hat{\pi}_k(t,i)=1$}{Sell asset at $t$\;}
{Hold asset at $t$\;}
}
}
\end{algorithm}
\paragraph{Deep Optimal Stopping.}
We use a deep learning-based optimal stopping benchmark (denoted  DOS for Deep Optimal Stopping) that is trained to learn the stopping policies given samples of the time series data and is now considered state-of-the-art in the optimal stopping literature~\cite{dos}. As in our framework, the authors decompose the problem of finding the best time to stop (and sell the asset) into a series of stopping decisions to be made at every time step (called the stopping policy $\pi(\cdot,\cdot)$ in our notation). They model this stopping policy at every time instant with a neural network (NN) that is fit to minimize the difference between the left and right sides of (\ref{eq:value_fn_dp}). To have a fair comparison between our algorithms and DOS, we retain the pre-processing part including clustering the training data before learning the stopping policy within each cluster. An important note is that DOS learns as many NNs as there are time steps $T$, so that the stopping policy at a time step is given by a different NN than that at other time steps.

\section{Experiments}\label{sec:expts}

In this section, we compare the performance of our algorithms GPOS and A-GPOS, against that of SOS and DOS using the suboptimality metric (\ref{eq:suboptimality})\footnote{The NN configuration for DOS was chosen as in section 4.1 in their paper \cite{dos} for the Bermudan max call option ($d = 1$, $I = 2$ and $q_1 = q_2 = 40+1$). For the DGPs, the choice was by trial and error to maximize predictive performance during training. We refrain from sharing code at the submission phase to retain anonymity.}. We also include performance metrics for Deep GP based Optimal Stopping (denoted DGPOS henceforth) which corresponds to fitting Deep GP models in line 4 of Algorithm \ref{alg:gpos}. Similarly, denote Adaptive Deep GP based Optimal Stopping by A-DGPOS reflecting fitting a Deep GP in line 6 of Algorithm \ref{alg:agpos}. Each algorithm is tried out on price series clustered using the DTW clustering algorithm described in Section \ref{subsec:clustering}.

\begin{algorithm}[tb]
\caption{A-GPOS: Adaptive Gaussian Process based Optimal Stopping\label{alg:agpos}}
\DontPrintSemicolon
\SetKwFunction{Training}{Training}
\SetKwProg{Fn}{Function}{:}{}
  \Fn{\Training{$\mathcal{D}$}}{
    Partition $\mathcal{D}$ into $K\geq1$ clusters $\mathcal{D}_1,\cdots,\mathcal{D}_K$\;
    \KwRet{$\mathcal{D}_1,\cdots,\mathcal{D}_K$}\;
  }
\SetKwFunction{Testing}{Adaptive Testing}
\SetKwProg{Fn}{Function}{:}{}
\Fn{\Testing{$\mathcal{D}_1,\cdots,\mathcal{D}_K$, $y^*$}}{
Find cluster $k$ for $y^*$\;
Fit GP or DGP to \emph{centroid} of cluster $k$ initialized with $\lbrace y^*_1,\cdots,y^*_W\rbrace$\;
Compute $\hat{V}_{t,i}(\mathcal{D}_k\cup y^*)$ as in (\ref{eq:gpos_value}) for all $t>W,i$\;
Compute $\hat{\pi}_k(t,i)$ as in (\ref{eq:gpos_policy}) for all $t>W,i$\;
\For{$t\in\lbrace W+1,\cdots,T\rbrace$}{
Find bin $b_i=[b_i^-,b_i^+]\ni y^*_t$\;
\eIf{$\hat{\pi}_k(t,i)=1$}{Sell asset at $t$\;}
{Hold asset at $t$\;}
}
}
\end{algorithm}

We consider four datasets including synthetic mean reverting time series generated from GPs, intra-day equity stock prices from early 2019, end-of-day equity stock prices from 2000 until 2020 and daily US treasury rates from 1990 until 2020. We do a 70-30 training and test split after randomly shuffling our normalized price data, and consider the stopping problem of finding the best time to sell the asset during a certain time horizon after observing its prices over the initial one third of the time horizon. 

\subsection{Synthetic Data Experiments}

We generate synthetic time series following an Ornstein–Uhlenbeck (OU) process which is a mean-reverting process popularly studied in financial literature \cite{nongaussianou,optionsou}. An OU process $\lbrace x_t:t\geq0\rbrace$ is a stochastic process with the following mean and covariance functions \begin{align}
    \mathbb{E}[x_t]&=x_0e^{-\theta t}+\mu\left(1-e^{-\theta t}\right)\nonumber\\
    \mathrm{Cov}[x_s,x_t]&=\frac{\sigma^2}{2\theta}\left(e^{-\theta|t-s|}-e^{-\theta|t+s|}\right)\nonumber
\end{align}
where $\theta>0$ is the mean reversion parameter, $\mu$ is a drift term and $\sigma>0$ is a variance parameter. OU processes are a special type of Gaussian processes, those with an exponential kernel \cite{gpml}. We generate samples of synthetic OU series from the same GP with different seeds. Figure \ref{fig:perf} (a) shows suboptimality values with each row corresponding to a different number of clusters $K$. We see that Adaptive GPOS with no clustering does the best, with GPOS and Deep GPOS doing the next best in suboptimality. As expected for synthetic data generated from a GP, algorithms using a GP model with the same kernel structure perform slightly better than those using DGP models. 

\subsection{Historical Price Data Experiments}
We look at three historical price datasets -- intra-day and end-of-day equity stock prices for nearly two dozen stocks, and 1 Yr US treasury rates. Since we have access to prices of more than one asset within each dataset, we devise a \emph{universal} model for prices across assets by fitting a GP to (centroids of) clusters of normalized prices from all assets. Intuitively, this would correspond to clustering assets with different characteristics such as volatility, into different clusters. 

\paragraph{Intra-day equity stock prices.}

\begin{figure*}[tb]
\begin{minipage}{0.49\linewidth}
\centering
\includegraphics[width=\linewidth]{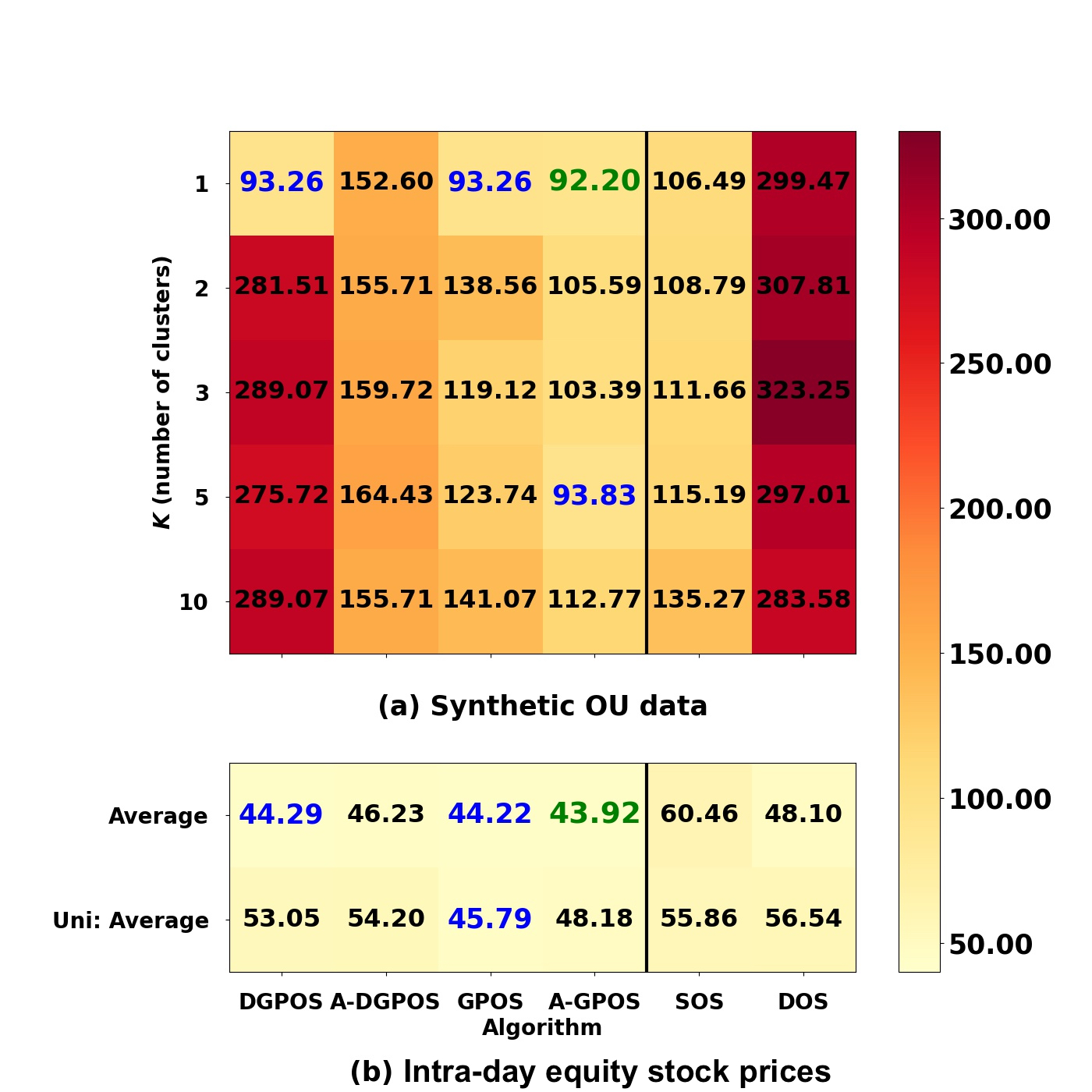}
\end{minipage}%
\begin{minipage}{0.49\linewidth}
\centering
\includegraphics[width=\linewidth]{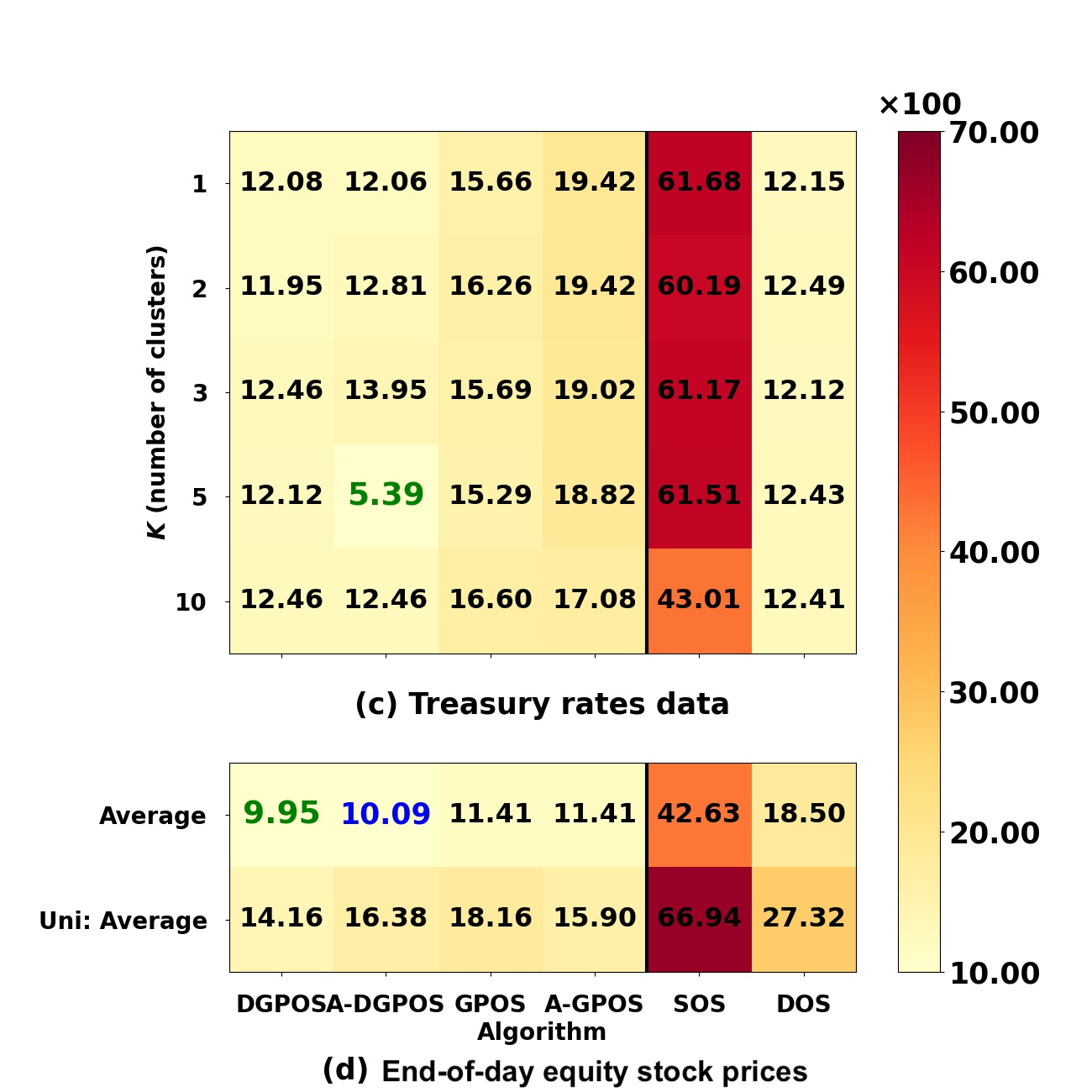}
\end{minipage}
\caption{Suboptimality in bps for four datasets. Every column specifies a stopping algorithm with a black vertical line separating our proposed algorithms from the benchmarks. Lower values of suboptimality corresponding to \colorbox{yellow}{yellow} regions are better than \colorbox{red}{red} regions. The least suboptimality (best performance) is shown in \textbf{\textcolor{mygreen}{bold green}}, while values within 5\% of it are shown in \textbf{\textcolor{myblue}{bold blue}}.}
\label{fig:perf}
\end{figure*}

We have intra-day equity stock prices measured every minute in a trading day for 26 out of the 30 underlying stocks of the Dow Jones Industrial Average (DJIA) in early 2019. The goal of the stopping problem is to find the best time to sell an asset during the trading day, given initial price data for the first one third of the day. Figure \ref{fig:perf} (b) is a matrix of suboptimality values with $K=3$ clusters where the first row displays the average suboptimality of stopping algorithms across the 26 stocks considered. We are able to compute the average across stocks using an arithmetic mean of suboptimality values for each stock since we express suboptimality in bps. And, the second row displays the average suboptimality of the universal model across all stocks. We see that our proposed algorithms beat the benchmarks with adaptive GPOS doing the best, and with DGP based algorithms being within 5\% of the best. 

\paragraph{End-of-day equity stock prices.}
We examine prices of 23 out of the 30 underlying stocks of the DJIA that traded from 2000 to 2020 at the end of every business day. The goal of the stopping problem is to find the best time to sell an asset during a year, given prices for the first one third of the year. Figure \ref{fig:perf} (d) shows suboptimality (divided by 100) with $K=3$ clusters where the first row displays average suboptimality of algorithms across all stocks. And, the second row displays average suboptimality of the universal model across all stocks. We see that our proposed algorithms beat the benchmarks with DGPOS doing the best. 

\begin{figure}[tb]
    \centering
    \includegraphics[width=\linewidth,height=1.7in]{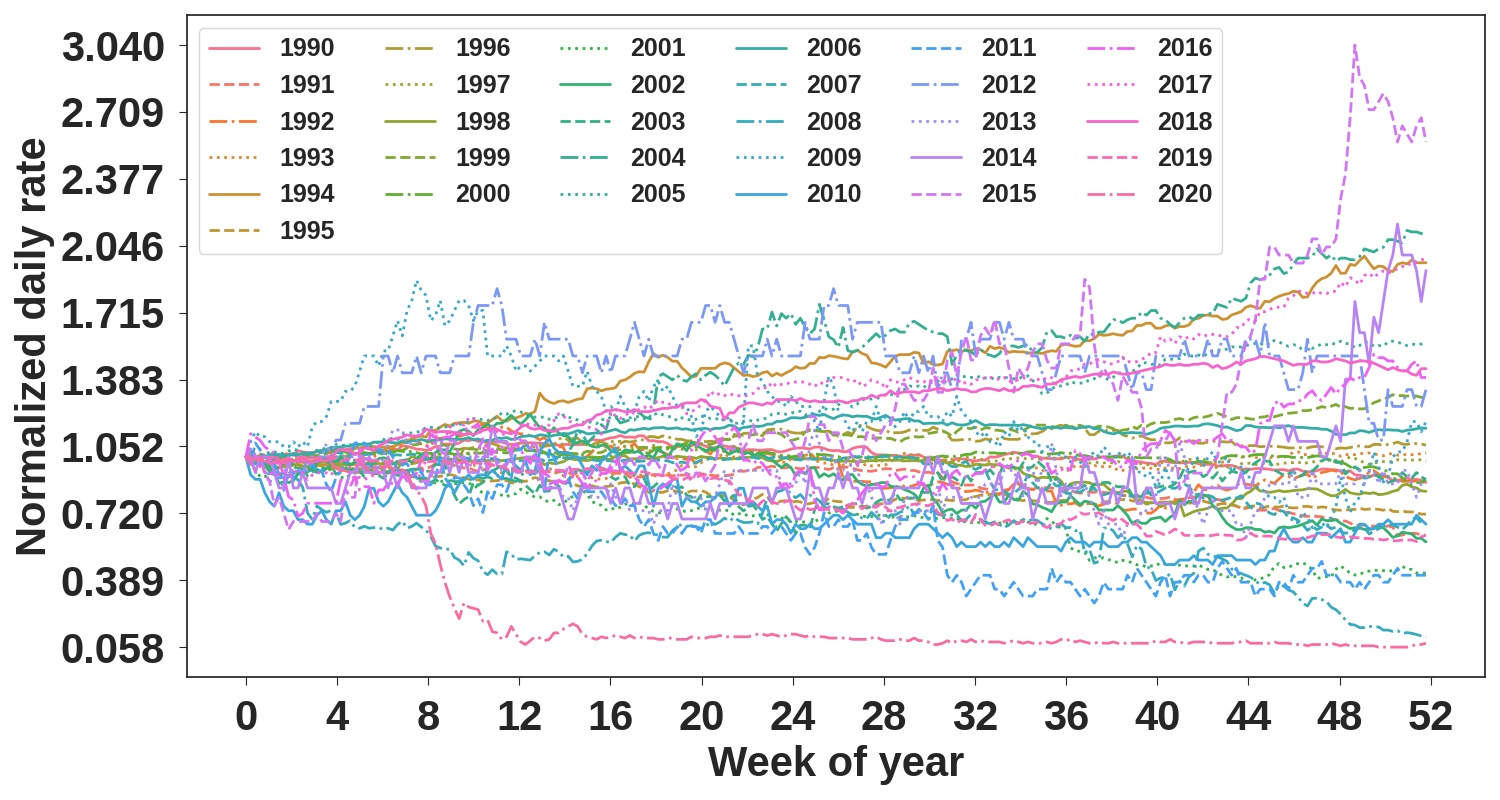}
    \caption{Normalized daily treasury rates from 1990 - 2020}
    \label{fig:tsy_rates_data}
\end{figure}

\paragraph{Treasury rates data.}
We look at daily yield curve rates for the 1Yr Treasury from 1990 to 2020 for the US Treasury publicly available at \cite{tsy_rates_data_source}. The goal of the stopping problem is to predict for the day in the year when rates would reach their maximum, given initial rates data for about one third of the year. Figure \ref{fig:tsy_rates_data} is a plot of the treasury rates normalized by their initial values in the year with each curve representing a year of data. We clearly observe trends in the rates data that require the use of a non-stationary kernel (one that depends on absolute time index) for a GP that is to be fit to this data\footnote{Notice the visible non-stationarity in year 2020 due to the COVID rate regulation!}. That is to say that a GP with a exponential kernel would perform poorly on the treasury rates data while doing well on the previous asset price data. Recollect that the DGP came up as the result of trying to overcome such a pre-defined kernel choice for GP models.  
Figure \ref{fig:perf} (c) shows the suboptimality of our proposed algorithms (divided by 100) with rows specifying the number of clusters $K$. As expected, we see that adaptive DGPOS performs better than GP based methods (without a specially picked kernel) while beating both benchmarks.

\section{Discussion and Conclusion}

We propose a family of algorithms to generate stopping rules for time series that can be modeled using Gaussian Processes. We utilize the Gaussian Process structure to design an analytical approach to compute stopping value functions (and their distributions) and stopping policies. We compare and contrast our family of algorithms with a sampling based benchmark as well as one from the deep learning literature. We demonstrate better performance over the benchmarks on three historical financial time series data sets. We also note that our algorithms are more computationally efficient than the benchmarks.




%

\begin{acks}
This paper was prepared for informational purposes by the Artificial Intelligence Research group of JPMorgan Chase \& Co. and its affiliates (``JP Morgan''), and is not a product of the Research Department of JP Morgan. JP Morgan makes no representation and warranty whatsoever and disclaims all liability, for the completeness, accuracy or reliability of the information contained herein. This document is not intended as investment research or investment advice, or a recommendation, offer or solicitation for the purchase or sale of any security, financial instrument, financial product or service, or to be used in any way for evaluating the merits of participating in any transaction, and shall not constitute a solicitation under any jurisdiction or to any person, if such solicitation under such jurisdiction or to such person would be unlawful.
\end{acks}

\printbibliography

\appendix

\end{document}